%% file: SchurVINS.tex
\documentclass[10pt,twocolumn,letterpaper]{article}
\usepackage[final]{cvpr}      

\input{preamble}

\definecolor{cvprblue}{rgb}{0.21,0.49,0.74}
\usepackage[pagebackref,breaklinks,colorlinks,citecolor=cvprblue]{hyperref}
\usepackage{rotating}
\usepackage{threeparttable}
\def\paperID{6796} 
\def\confName{CVPR}
\def\confYear{2024}

\title{SchurVINS: Schur Complement-Based Lightweight Visual Inertial Navigation System}

\author{Yunfei Fan, Tianyu Zhao, Guidong Wang\\
ByteDance\\
{\tt\small \{frank.01, zhaotianyu.1998, guidong.wang\}@bytedance.com}
}

\begin{document}
\maketitle
\input{sec/0_abstract}    
\input{sec/1_intro}

\input{sec/2_related_work}
\input{sec/3_framework}
\input{sec/4_experiment}
\input{sec/5_conclusion}

\input{sec/6_acknowledgment}

{
    \small
    \bibliographystyle{ieeenat_fullname}
    \bibliography{SchurVINS}
}


\end{document}

%% file: preamble.tex
\usepackage[dvipsnames]{xcolor}


%% file: sec/0_abstract.tex
\begin{abstract}
Accuracy and computational efficiency are the most important metrics to Visual Inertial Navigation System (VINS). The existing VINS algorithms with either high accuracy or low computational complexity, are difficult to provide the high precision localization in resource-constrained devices. To this end, we propose a novel filter-based VINS framework named \textbf{S}chur\textbf{V}INS (SV), which could guarantee both high accuracy by building a complete residual model and low computational complexity with Schur complement. Technically, we first formulate the full residual model where Gradient, Hessian and observation covariance are explicitly modeled. Then Schur complement is employed to decompose the full model into ego-motion residual model and landmark residual model. Finally, Extended Kalman Filter (EKF) update is implemented in these two models with high efficiency. Experiments on EuRoC and TUM-VI datasets show that our method notably outperforms state-of-the-art (SOTA) methods in both accuracy and computational complexity. The experimental code of SchurVINS is available at \url{https://github.com/bytedance/SchurVINS}.
\end{abstract}

%% file: sec/1_intro.tex
\section{Introduction}
\label{sec:intro}

High-precision localization technologies have become a cornerstone in various industrial fields, playing an indispensable role particularly in robotics, augmented reality (AR), and virtual reality (VR). In recent decades, visual inertial navigation system (VINS) has attracted significant attentions due to its advantages of low-cost and ubiquitousness. Composed of only cameras and inertial measurement units (IMU), the VINS module can provide six-degree-of-freedom (6-DOF) positioning as accurate as expensive sensors such as Lidar, and is more competent in being installed in portable devices like smartphone and micro aerial vehicles (MAV).

\begin{figure}[t]
  \centering
  \includegraphics[width=\linewidth]{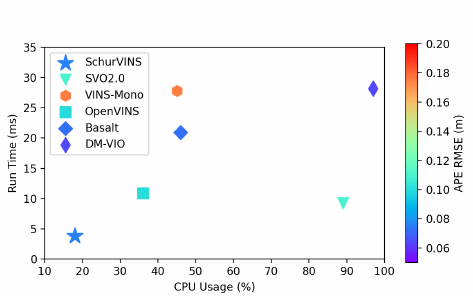}
   \caption{Comparison of run time, CPU usage and RMSE evaluated on EuRoC dataset. Different shapes and colors indicate different methods and precision, respectively.}
   \label{fig:best_performance}
\end{figure}

It has been reported that kinds of excellent open-source VINS algorithms could achieve high-precision pose estimation, which mainly includes two methodologies: optimization-based and filter-based methods. Typical optimization-based methods~\cite{okvis_2015keyframe,qin2018vinsmono,liu2018iceba,basalt2019,orbslam3_2021,dmvio2022} model poses and the corresponding observed landmarks jointly. Benefitting from Schur complement technique~\cite{schur_complement}, this high-dimensional model with special sparsity could be solved efficiently by bundle adjustment (BA~\cite{ba2000}). In theory~\cite{concise_review_2019huang}, although notable in high-precision of localization, optimization-based methods may suffer from high computational complexity. In contrast, main-stream filter-based methods~\cite{rovio2017iterated,sun2018smsckf,geneva2020openvins,fan2020smsckf} derived from MSCKF~\cite{mourikis2007msckf} utilize the left nullspace method to simplify the residual model. EKF~\cite{sola2017eskf} update is then executed on the simplified residual model to estimate corresponding poses. Finally, they achieve high efficiency but compromise accuracy, since landmarks are not optimized with camera poses jointly and all observations are utilized only once. To sum up, optimization-based methods are advantageous in accuracy while filter-based methods are more efficient.

Therefore, it is urgent to develop a framework combines their high precision and efficiency. As discussed above, traditional residual model without simplification can achieve high accuracy. In spite of this, when both landmarks and poses are incorporated into the state vector for joint estimation, the efficiency of EKF-SLAM significantly decreases~\cite{mourikis2007msckf}. Inspired by the Schur comple-ment in optimization-based methods, we make full use of the sparse structure inherent in the high-dimensional residual model constructed with poses and landmarks to achieve high efficiency in EKF. Thus, an EKF-based VINS framework that achieves both high efficiency and accuracy is presented. In the framework, the equivalent residual model, consisting of gradient, Hessian and the corresponding observation covariance, is derived based on the traditional residual model. Taking the special sparse structure of Hessian into account, Schur complement is carried out to break the equivalent residual equation into two smaller equations: equivalent pose residual model and equivalent landmark residual model. The equivalent landmark residual model is able to be further split into a collection of small equivalent residual models due to its own sparse structure. Finally, EKF update is implemented with the derived equivalent residual model to estimate the poses and corresponding landmarks jointly. As shown in ~\cref{fig:best_performance}, the resulting framework outperforms SOTA methods in latency, computational complexity and accuracy. Our main contributions are summarized as follows:
\begin{itemize}
  \item An equivalent residual model is proposed to deal with hyper high-dimension observations, which consists of gradient, Hessian and the corresponding observation covariance. This method is of great generality in EKF systems.
  \item A lightweight EKF-based landmark solver is proposed to estimate position of landmarks with high efficiency.
  \item A novel EKF-based VINS framework is developed to achieve ego-motion and landmark estimation simultaneously with high accuracy and efficiency. The experimental code is published to benefit community.
  \end{itemize}

  \begin{figure*}[tbhp]
    \centering
    \includegraphics[width=\linewidth]{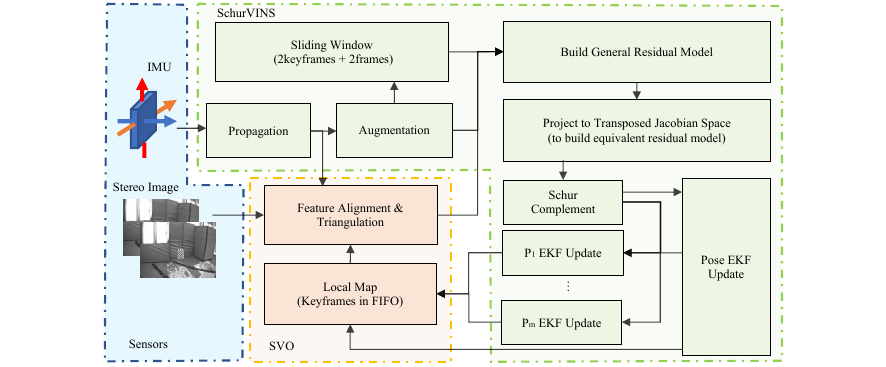}
     \caption{Framework of SchurVINS, which shows the relationship between SVO and SchurVINS. $P_1$ to $P_m$ represent the valid landmarks of the surrounding environment which are employed to construct residual model.}
     \label{fig:FullFrame_Fig1}
 \end{figure*}

%% file: sec/2_related_work.tex
\section{Related Work}
\label{sec:Related_Work}
Improving the efficiency and accuracy is an ongoing effort for VINS algorithms. To date, significant research has been carried out to reduce the computational complexity and improve the precision. 

Many VINS algorithms focus on efficiency improve-ment. Some studies reuse the intermediate results of previous optimization to decrease the amount of repetitive computation~\cite{isam2008isam,isam2_2012kaess,slam++2017slam,liu2018iceba}. While these approaches may yield a slight loss in accuracy, the computational process can be notably accelerated. Some other studies try to achieve high efficiency through engineering technologies. In ~\cite{polok2013efficient,ye2022coli}, efficient Hessian construction and Schur complement calculation is employed to improve cache efficiency and avoid redundant matrix representation. In ~\cite{wkj2015square,squareroot2021}, variables are declared by single precision instead of traditional double precision to speed up the algorithm.

Besides efficiency, some studies concentrate on improving the accuracy. In ~\cite{fej2009,huang2008analysis,msckf2_2013high}, high accuracy is guaranteed through improving the consistency in EKF-based VINS. Some improved MSCKF namely Hybrid MSCKF~\cite{li_hybrid_2012vision,geneva2020openvins} (combined MSCKF and EKF-SLAM), proposed in recent to balance efficiency and accuracy, model informative landmarks selectively as part of their state variables to estimate jointly~\cite{li2013optimization}. Some researchers construct the local bundle adjustment (LBA) running on other threads to reduce drift~\cite{Forster17troSVO,orbslam3_2021}. However, LBA requires massive computational resources which might not be practical for implementation on small devices.

%% file: sec/3_framework.tex
\section{SchurVINS Framework}
\label{sec:Framework}

In this paper, the proposed SchurVINS is developed based on open-source SVO2.0~\cite{Forster14icra,Forster17troSVO} with stereo configuration, in which sliding window based EKF back-end is employed to replace the original back-end in SVO2.0, and EKF-based landmark solver is utilized to replace the original landmark optimizer. The framework of SchurVINS algorithm and the relationship between SVO and SchurVINS are shown in \cref{fig:FullFrame_Fig1}. 

\subsection{State Definition} 

Normally, for a traditional EKF-based VINS system~\cite{geneva2020openvins,msckf2_2013high,fan2020smsckf}, the basic IMU state is defined as:

\begin{equation}
\textbf{x}_I=
\begin{bmatrix}
{^G_I\textbf{q}}^\mathsf{T}&{^G\textbf{p}_I}^\mathsf{T}&{^G\textbf{v}_I}^\mathsf{T}&\textbf{b}_a^\mathsf{T}&\textbf{b}_g^\mathsf{T}
\end{bmatrix}^\mathsf{T}
\label{eqn-1}
\end{equation}
where \{G\}, \{I\} and \{C\} are the global frame, local frame and camera frame, respectively. ${^G\textbf{p}_I}$ and ${^G\textbf{v}_I}$ are the position and velocity of IMU expressed in \{G\}, respectively. ${^G_I\textbf{q}}$ represents the  rotation quaternion from \{I\} to \{G\} (in this paper, quaternion obeys Hamilton rules~\cite{sola2017eskf}). The vectors $\textbf{b}_a$ and $\textbf{b}_g$ individually represent the biases of the angular velocity and linear acceleration measured by the IMU device. And thus the corresponding EKF error-state of ${\textbf{x}_I}$ is defined as \cref{eqn-2}
\begin{equation}
\tilde{\textbf{x}}_I=
\begin{bmatrix}
{^{G}_{I}\tilde{\boldsymbol{\theta}}}^\mathsf{T}&{^G{\tilde{\textbf{p}}}_I}^\mathsf{T}&{^G{\tilde{\textbf{v}}}_I}^\mathsf{T}&{{\tilde{\textbf{b}_a}}}^\mathsf{T}&{{\tilde{\textbf{b}_g}}}^\mathsf{T}
\end{bmatrix}^\mathsf{T}
\label{eqn-2}
\end{equation}
where, ${^{G}_{I}\tilde{\boldsymbol{\theta}}}$ represents the error-state of $^{G}_{I}\textbf{q}$. Except for quaternion, other states can be used with standard additive error (e.g. $\textbf{x} = \hat{\textbf{x}} + \tilde{\textbf{x}}$). Similar to \cite{sola2017eskf}, the extended additive error of quaternion is defined as \cref{eqn-3}  (in this paper, quaternion error is defined in frame $\{G\}$) 
\begin{equation}
\begin{matrix}
   \textbf{q}^G_I={\delta{^G_I\textbf{q}}} \otimes {^G_I\hat{\textbf{q}}}, & 
   {\delta{^G_I\textbf{q}}} = 
   \begin{bmatrix}
      1 & \frac{1}{2}\delta{^G_I{\tilde{\boldsymbol{\theta}}}}
   \end{bmatrix}^\mathsf{T}
\end{matrix}
\label{eqn-3}
\end{equation}

Similarly, the extended additive error of rotation matrix is defined as \cref{eqn-4}
\begin{equation}
\begin{matrix}
   \textbf{R}({^G_I\textbf{q}}) = {^G_I\textbf{R}}, &
   {^G_I\textbf{R}} = 
   \begin{pmatrix} 
      \textbf{I} + 
      \begin{bmatrix}
         ^G_I\tilde{\boldsymbol{\theta}}
      \end{bmatrix}_{\times}      
   \end{pmatrix} {^G_I\hat{\textbf{R}}}

\end{matrix}
\label{eqn-4}
\end{equation}

\subsection{Propagation and Augmentation}

SchurVINS follows the policy introduced in~\cite{sola2017eskf} on state propagation. The time evolution of IMU states are described as
\begin{gather}
\begin{matrix}
^G_I\dot{\hat{\textbf{q}}}= \frac{1}{2}{^G_I{\hat{\textbf{q}}}}\otimes \Omega(\hat{\boldsymbol{\omega}}),&
\Omega{(\hat{\boldsymbol{\omega}})}=
\begin{pmatrix}
   0&-{\hat{\boldsymbol{\omega}}}^\mathsf{T} \\
   \hat{\boldsymbol{\omega}}&-[\hat{\boldsymbol{\omega}}]_\times
\end{pmatrix}
\end{matrix} 
\label{eqn-5}\\
\begin{matrix}
   \dot{\hat{\textbf{b}}}_g= \textbf{0}_{3\times1},&\dot{\hat{\textbf{b}}}_a=\textbf{0}_{3\times1}
\end{matrix} 
\label{eqn-6}\\
\begin{matrix}
^G{\dot{\hat{\textbf{p}}}}_I={^G\textbf{v}_I},&^G\dot{\hat{\textbf{v}}}_I={^G_I{\hat{\textbf{R}}}}{\hat{\textbf{a}}}+{^G\textbf{g}}
\end{matrix}
\label{eqn-7}
\end{gather}
where $\hat{\boldsymbol{\omega}}=\boldsymbol{\omega}_m-\hat{\textbf{b}}_g$ and $\hat{\textbf{a}}=\textbf{a}_m-\hat{\textbf{b}}_a$ are IMU measurements with biases discarded.
where $[\hat{\boldsymbol{\omega}}]_\times$ is skew symmetric matrix of $\hat{\boldsymbol{\omega}}$. Based on Eqs.\eqref{eqn-5} to~\eqref{eqn-7}, the linearized continuous dynamics for the error IMU state is defined as
\begin{equation}
\dot{\tilde{\textbf{x}}}_I=\textbf{F}\tilde{\textbf{x}}_I + \textbf{G}\textbf{n}_I
\label{eqn-9}
\end{equation}
where ${\textbf{n}_I}=\begin{bmatrix}
   {\textbf{n}_a}^\mathsf{T}&{\textbf{n}_{a\omega}}^\mathsf{T}&{\textbf{n}_{g}}^\mathsf{T}&{\textbf{n}_{g\omega}}^\mathsf{T}
\end{bmatrix}^\mathsf{T}$. Vectors $\textbf{n}_a$ and $\textbf{n}_g$ represent the Gaussian noise of the accelerometer and gyroscope measurement, while $\textbf{n}_{a\omega}$ and $\textbf{n}_{g\omega}$ are the random walk rate of the accelerometer and gyroscope measurement biases. $\textbf{F}$ and $\textbf{G}$ are defined as
\begin{align}
\textbf{F}&=\begin{bmatrix}
\textbf{0}_{3\times3}&\textbf{0}_{3\times3}&\textbf{0}_{3\times3}&\textbf{0}_{3\times3}&{-{^G_I\textbf{R}}} \\
\textbf{0}_{3\times3}&\textbf{0}_{3\times3}&\textbf{I}_{3\times3}&\textbf{0}_{3\times3}&\textbf{0}_{3\times3} \\
{-[{^G_I\textbf{R}}{\hat{\textbf{a}}}]_{\times}}&\textbf{0}_{3\times3}&\textbf{0}_{3\times3}&{-{^G_I\textbf{R}}}&\textbf{0}_{3\times3} \\
\textbf{0}_{6\times3}&\textbf{0}_{6\times3}&\textbf{0}_{6\times3}&\textbf{0}_{6\times3}&\textbf{0}_{6\times3}&
\end{bmatrix}     \label{eqn-10}     \\
\textbf{G}&=\begin{bmatrix}
\textbf{0}_{3\times3}&\textbf{0}_{3\times3}&{-{^G_I\textbf{R}_{3\times3}}}&\textbf{0}_{3\times3} \\
\textbf{0}_{3\times3}&\textbf{0}_{3\times3}&\textbf{0}_{3\times3}&\textbf{0}_{3\times3} \\
-{^G_I\textbf{R}_{3\times3}}&\textbf{0}_{3\times3}&\textbf{0}_{3\times3}&\textbf{0}_{3\times3} \\
\textbf{0}_{3\times3}&\textbf{I}_{3\times3}&\textbf{0}_{3\times3}&\textbf{0}_{3\times3} \\
\textbf{0}_{3\times3}&\textbf{0}_{3\times3}&\textbf{0}_{3\times3}&\textbf{I}_{3\times3} \\
\end{bmatrix}
\label{eqn-11}
\end{align}

$4^{th}$ Runge-Kutta numerical integration method is employed in Eqs.~\eqref{eqn-3} to~\eqref{eqn-7} for propagating the estimated IMU state. Based on~\cref{eqn-9}, the discrete time state transition matrix $\boldsymbol{\Phi}$ and discrete time noise covariance $\textbf{Q}$ are formulated as follows:
\begin{equation}
\begin{aligned} 
{\boldsymbol{\Phi}}&=\textbf{I}_{15\times15} + \textbf{F}dt + \frac{1}{2}\textbf{F}^2dt^2 + \frac{1}{6}\textbf{F}^3dt^3 \\
\textbf{Q}&={\boldsymbol{\Phi}}\textbf{G}\textbf{Q}_I\textbf{G}^\mathsf{T}{\boldsymbol{\Phi}}^\mathsf{T}dt
\end{aligned}
\label{eqn-12}
\end{equation}
where $\textbf{Q}_I=E[\textbf{n}_I{\textbf{n}_I}^\mathsf{T}]$ is the continuous time noise covariance matrix of the system. Hence, the formulations of covariance propagation are built as:
\begin{equation}
   \begin{matrix}
    \textbf{P}_{II} \leftarrow {\boldsymbol{\Phi}}\textbf{P}_{II}{\boldsymbol{\Phi}}^\mathsf{T} + \textbf{Q}, &\textbf{{P}}_{IA} \leftarrow {\boldsymbol{\Phi}}\textbf{P}_{IA}
   \end{matrix}
\label{eqn-13}
\end{equation}

The covariance $\textbf{P}$ is partitioned as ~\cref{eqn-14}. $\textbf{P}_{II}$ is the covariance of basic state. $\textbf{P}_{IA}$ and $\textbf{P}_{AI}$ is the covariance between basic state and augmented state. $\textbf{P}_{AA}$ is covariance of the augmented state.
\begin{equation}
   \textbf{P}=\begin{bmatrix}
      \textbf{P}_{II}&\textbf{P}_{IA}\\
      \textbf{P}_{IA}^\mathsf{T}&\textbf{P}_{AA}\\
   \end{bmatrix}  
\label{eqn-14} 
\end{equation}

When a new image arrives, the current IMU pose $\textbf{x}_{Ai}=
\begin{bmatrix}
   {^G_I\textbf{q}}^\mathsf{T}&{^G\textbf{p}_I}^\mathsf{T}
\end{bmatrix}^\mathsf{T}$ is augmented as well as its covariance. The augmentation formulations are:
\begin{equation}
\begin{aligned}
\textbf{X}&=\begin{bmatrix}
    {\textbf{x}_I}^\mathsf{T}&{\textbf{x}_{A0}}^\mathsf{T}&{\textbf{x}_{A1}}^\mathsf{T}&{\dotsb}&{\textbf{x}_{Ai}}^\mathsf{T}
 \end{bmatrix}^\mathsf{T}            \\
\textbf{P} &\leftarrow \begin{bmatrix}
    \textbf{P}&{\textbf{P}_{21}}^\mathsf{T} \\
    \textbf{P}_{21}&\textbf{P}_{22}
 \end{bmatrix}
\end{aligned}
\label{eqn-15}
\end{equation}
where $\textbf{P}_{21}=\textbf{J}_a\textbf{P}$, $\textbf{P}_{22}=\textbf{J}_a\textbf{P}{\textbf{J}_a}^\mathsf{T}$. And $\textbf{J}_a$ is the Jacobian of $\tilde{\textbf{x}}_{Ai}$ with respect to error states, which is defined as follows:
\begin{equation}
\textbf{J}=\begin{bmatrix}
    \textbf{I}_{3\times3}&\textbf{0}_{3\times3}&\textbf{0}_{3\times(9+6N)}\\
    \textbf{0}_{3\times3}&\textbf{I}_{3\times3}&\textbf{0}_{3\times(9+6N)}\\
\end{bmatrix}
\label{eqn-16}
\end{equation}

\begin{figure}[tbp]
   \centering
   \includegraphics[width=\linewidth]{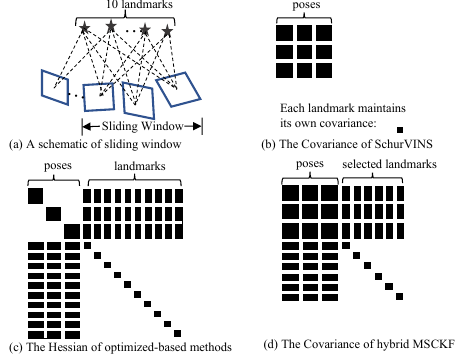}
    \caption{A schematic of our system for ten landmarks and the sliding window of size three shown in (a), and the Hessian or Covariance of different methods shown in (b)-(d). (b) shows our algorithm in which the covariance of every single landmark is independent from the entire covariance of poses in the sliding window. (c) demonstrates the Hessian of both landmarks and poses in the sliding window. (d) demonstrates traditional hybrid MSCKF with the Covariance of both selected landmarks and poses in the sliding window.}
    \label{fig:CovAndHessianCompareFig2}
\end{figure}
\subsection{Schur Complement-Based State Update}
In the SchurVINS scheme, unlike MSCKF methods~\cite{geneva2020openvins,sun2018smsckf}, the EKF update is conducted based on all the successfully triangulated landmarks and their observations in the sliding window, which can eliminate the drift caused by state propagation in every single image timestamp interval as much as possible. For single observation, the reprojection error $\textbf{r}_{i,j}$ of camera measurement is formulated as:
\begin{equation}
\begin{aligned}
   \textbf{r}_{i,j}&= {\textbf{z}}_{i,j} - \hat{\textbf{z}}_{i,j}  \\
   \textbf{r}_{i,j}&=\textbf{J}_{x,i,j}{\tilde{\textbf{X}}}+\textbf{J}_{f,i,j}{^G{\tilde{\textbf{p}}_{f_j}}}+\textbf{n}_{i,j}  \\
\hat{\textbf{z}}_{i,j}&=\frac{1}{{^{C_i}\hat{Z}_j}}
\begin{bmatrix}
   {^{C_i}\hat{X}_j} \\
   {^{C_i}\hat{Y}_j}
\end{bmatrix}
\end{aligned}
\label{eqn-17}
\end{equation}
where $\textbf{r}_{i,j}$ and $\textbf{z}_{i,j}$ are the reprojection error and the camera measurement of $j^{th}$ landmark at $i^{th}$ pose in sliding window, respectively, and ${\hat{\textbf{z}}_{i,j}}$ is the corresponding theoretical measurement formulated by estimated states. 
${^{C_{i}}\textbf{p}_j}=\begin{bmatrix}
   {^{C_i}\hat{X}_j}&{^{C_i}\hat{Y}_j}&{^{C_i}\hat{Z}_j}
\end{bmatrix}$ is the landmark coordinate in camera pose of $i^{th}$ sliding window.
$\textbf{n}_{i,j}$ represents the corresponding measurement noise. 
$\tilde{\textbf{X}}$ and ${^G{\tilde{\textbf{p}}}_{f_j}}$ are respectively the state perturbation and landmark position perturbation.
$\textbf{J}_{x,i,j}$ and $\textbf{J}_{f,i,j}$ are the Jacobians of residual with respect to system state and landmark position, respectively. 
The Jacobians are defined as follows:
\begin{equation}
\begin{aligned}
   \textbf{J}_{x,i,j} &= 
      \begin{bmatrix}
         \textbf{0}_{2\times{(15+6i)}}&\textbf{J}_A&\textbf{0}_{2\times{6(N-i-1)}}
      \end{bmatrix}                 \\
      \textbf{J}_A&=\textbf{J}_{i,j}\begin{bmatrix}
         {^I_C{\hat{\textbf{R}}^\mathsf{T}}}[{^{I_i}\hat{\textbf{p}}_{f_j}}]_{\times}{^G_{I_i}\hat{\textbf{R}}^\mathsf{T}}  &-{^G_{C_i}{\hat{\textbf{R}}^\mathsf{T}}}
      \end{bmatrix}                  \\
      \textbf{J}_{f,i,j}&=\begin{bmatrix}
         \textbf{J}_{i,j}{^G_{C_i}{\hat{\textbf{R}}^\mathsf{T}}}   
      \end{bmatrix}
\end{aligned}
\label{eqn-18}
\end{equation}
where, for convenience, we define the camera model using the pinhole model. Therefore, $\textbf{J}_{i,j}$ is defined as:

\begin{equation}
   \textbf{J}_{i,j}=
   \frac{1}{{^{C_i}{\hat{Z}}_j}^2}
   \begin{bmatrix}
   {^{C_i}{\hat{Z}}_j}&0&-{^{C_i}{\hat{X}}_j}\\
   0&{^{C_i}{\hat{Z}}_j}&-{^{C_i}{\hat{Y}}_j}\\
\end{bmatrix}
\label{eqn-19}
\end{equation}

Aiming at all the observations of landmarks in the sliding window, we can acquire the full residual model by stacking all the residual equations:
\begin{equation}
\begin{matrix}
   \textbf{r}=\begin{bmatrix}
      \textbf{J}_x&\textbf{J}_f
    \end{bmatrix}
    \begin{bmatrix}
        {\tilde{\textbf{X}}} \\
        {^G{\tilde{\textbf{p}}_f}}
    \end{bmatrix}+\textbf{n}
\end{matrix}
\label{eqn-20}
\end{equation}
where, $\textbf{r}$ and $\begin{bmatrix}
   \textbf{J}_x&\textbf{J}_f
 \end{bmatrix}$ are respectively the stacked residual and stacked Jacobian. $\textbf{J}_x$ and $\textbf{J}_f$ are jacobian with respect to states and landmark positions, respectively. $\textbf{n}$ is the stacked measurement noise, and the measurement covariance of $\textbf{n}$ is $\textbf{R}=diag(u^2, u^2, \dotsb, u^2)$, where $u$ is the element of standard deviation of $\textbf{n}$.
 
 Unlike~\cite{sun2018smsckf,geneva2020openvins,fan2020smsckf}, in this paper, the residual model~\cref{eqn-20} is projected into the jacobian space $\begin{bmatrix}
   \textbf{J}_x&\textbf{J}_f
 \end{bmatrix}^\mathsf{T}$for formulating equivalent residual equations, which consist of gradient and hessian and observation covariance shown in Eqs.~\eqref{eqn-21} and~\eqref{eqn-22} below. It is worth highlighting that this strategy is an alternative to QR decomposition strategy~\cite{mourikis2007msckf} for speeding-up in any EKF systems with high-dimensional measurements.
\begin{align}
    \begin{bmatrix}
    {\textbf{J}_x}^\mathsf{T} \\
     {\textbf{J}_f}^\mathsf{T}
    \end{bmatrix}
    \textbf{r}&=
    \begin{bmatrix}
    {\textbf{J}_x}^\mathsf{T} \\
     {\textbf{J}_f}^\mathsf{T}
    \end{bmatrix}
    \begin{bmatrix}
      \textbf{J}_x&\textbf{J}_f
    \end{bmatrix}
    \begin{bmatrix}
    \tilde{\textbf{X}} \\
     {^G\tilde{\textbf{P}}_f}
    \end{bmatrix}
    +\textbf{n}'          \label{eqn-21}                  \\
    \textbf{R}' &=
    \begin{bmatrix}
    {\textbf{J}_x}^\mathsf{T} \\
     {\textbf{J}_f}^\mathsf{T}
    \end{bmatrix}
    \textbf{R}\begin{bmatrix}
      \textbf{J}_x&\textbf{J}_f
    \end{bmatrix}   \label{eqn-22} 
\end{align}
where $\textbf{n}'$ and $\textbf{R}'$ are the equivalent observation noise and covariance, respectively. Obviously, Eqs.~\eqref{eqn-21} and~\eqref{eqn-22} could be simplified as:
\begin{figure}[t]
   \centering
   \includegraphics[width=\linewidth]{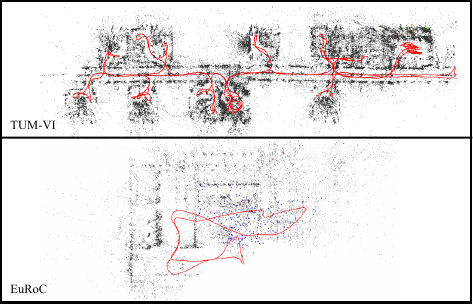}
    \caption{The experimental trajectory and point cloud of SchurVINS on TUM-VI and EuRoC datasets.}
    \label{fig:trajectory_and_points}
\end{figure}
\begin{align}
   \underbrace{
   \begin{bmatrix}
   {\textbf{J}_x}^\mathsf{T}\textbf{r} \\
    {\textbf{J}_f}^\mathsf{T}\textbf{r}
   \end{bmatrix}
   }_{
   \begin{bmatrix}
      \textbf{b}_1 \\ \textbf{b}_2
   \end{bmatrix}
   } &=
   \underbrace{
   \begin{bmatrix}
   {\textbf{J}_x}^\mathsf{T}\textbf{J}_x & {\textbf{J}_x}^\mathsf{T}\textbf{J}_f \\
   {\textbf{J}_f}^\mathsf{T}\textbf{J}_x & {\textbf{J}_f}^\mathsf{T}\textbf{J}_f
   \end{bmatrix}
   }_{
   \begin{bmatrix}
   {\textbf{C}_1} & \textbf{C}_2 \\
   {\textbf{C}_2}^\mathsf{T} & \textbf{C}_3
   \end{bmatrix}
   }
   \begin{bmatrix}
   \tilde{\textbf{X}} \\
    {^W\tilde{\textbf{P}}_f}
   \end{bmatrix}+
   \underbrace{\textbf{n}'}_{
      \begin{bmatrix}
         \textbf{n}'_1 \\ \textbf{n}'_2
      \end{bmatrix}
      }             \label{eqn-23}                 \\
      \textbf{R}' &=
   \begin{bmatrix}
   {\textbf{J}_x}^\mathsf{T}\textbf{J}_x & {\textbf{J}_x}^\mathsf{T}\textbf{J}_f \\
   {\textbf{J}_f}^\mathsf{T}\textbf{J}_x & {\textbf{J}_f}^\mathsf{T}\textbf{J}_f
   \end{bmatrix}u^2      \label{eqn-24}
\end{align}

Since $^G\tilde{\textbf{P}}_f$ is not included in the states in~\cref{eqn-15}, it is necessary to employ Schur complement~\cite{sibley2010sliding} on Eqs.~\eqref{eqn-21} and~\eqref{eqn-22} to marginalize the implicit states. To be straightforward, Eqs.~\eqref{eqn-23} and~\eqref{eqn-24} should be projected into $\textbf{L}$ space as Eqs.~\eqref{eqn-25} and~\eqref{eqn-26}.
\begin{align}
   \textbf{L}
   \begin{bmatrix}
      \textbf{b}_1 \\ \textbf{b}_2
   \end{bmatrix} &=
   \textbf{L}
   \begin{bmatrix}
      \textbf{C}_1 & \textbf{C}_2 \\
   {\textbf{C}_2}^\mathsf{T} & \textbf{C}_3
   \end{bmatrix}
   \begin{bmatrix}
   \tilde{\textbf{X}} \\
    {^W\tilde{\textbf{P}}_f}
   \end{bmatrix}+
   \begin{bmatrix}
      \textbf{n}''_1 \\
      \textbf{n}''_2
   \end{bmatrix} \label{eqn-25}     \\
   \textbf{R}'' &=\textbf{L}
   \begin{bmatrix}
      \textbf{C}_1 & \textbf{C}_2 \\
   {\textbf{C}_2}^\mathsf{T} & \textbf{C}_3
   \end{bmatrix}
   \textbf{L}^\mathsf{T}u^2=
   \begin{bmatrix}
      \textbf{R}_1''& \textbf{0} \\
      \textbf{0}&\textbf{R}_2''
   \end{bmatrix}
   \label{eqn-26}
\end{align}
where $
\begin{bmatrix}
   \textbf{n}_1''^{\mathsf{T}} & \textbf{n}_2''^{\mathsf{T}}
\end{bmatrix}^{\mathsf{T}}$ and $\textbf{R}''$ are the derived observation noise and covariance. And $\textbf{L}$ is defined as: 

\begin{equation}
   \textbf{L}=\begin{bmatrix}
      \textbf{I} & -\textbf{C}_2\textbf{C}_3^{-1} \\
      \textbf{0} & \textbf{I}
   \end{bmatrix} \label{eqn-27}
\end{equation}

Substituting~\cref{eqn-27} into Eqs.~\eqref{eqn-25} and~\eqref{eqn-26} yields the simplified formulations:
\begin{gather}
   \begin{bmatrix}
      \textbf{b}_1 - \textbf{C}_2\textbf{C}_3^{-1}\textbf{b}_2 \\ 
      \textbf{b}_2
\end{bmatrix}
   =\textbf{C}
   \begin{bmatrix}
   \tilde{\textbf{X}} \\
    {^W\tilde{\textbf{P}}_f}
   \end{bmatrix}+
   \begin{bmatrix}
      \textbf{n}''_1 \\ 
      \textbf{n}''_2
   \end{bmatrix}    \label{eqn-28}         \\
   \textbf{R}'' =\begin{bmatrix}
      (\textbf{C}_1-\textbf{C}_2\textbf{C}_3^{-1}\textbf{C}_2^\mathsf{T}) & \textbf{0} \\
      \textbf{0} & \textbf{C}_3
      \end{bmatrix} u^2     \label{eqn-29}
\end{gather}
where 
\begin{equation}
   \textbf{\textbf{C}}=\begin{bmatrix}
      (\textbf{C}_1-\textbf{C}_2\textbf{C}_3^{-1}\textbf{C}_2^\mathsf{T}) & \textbf{0} \\
      \textbf{C}_2^\mathsf{T} & \textbf{C}_3
      \end{bmatrix} 
      \label{eqn-29.1}
\end{equation}
Eqs.~\eqref{eqn-28} and~\eqref{eqn-29} could be decomposed into Eqs.~\eqref{eqn-30} to~\eqref{eqn-31} and Eqs.~\eqref{eqn-32} to~\eqref{eqn-33} as follows:
\begin{align}
\begin{bmatrix}
   \textbf{b}_1 - \textbf{C}_2\textbf{C}_3^{-1}\textbf{b}_2
    \end{bmatrix}
    &=
    \begin{bmatrix}
      \textbf{C}_1-\textbf{C}_2\textbf{C}_3^{-1}\textbf{C}_2^\mathsf{T}
    \end{bmatrix}
    \tilde{\textbf{X}}+\textbf{n}_1''      \label{eqn-30} \\
    \textbf{R}_1''&=\begin{bmatrix}
      \textbf{C}_1-\textbf{C}_2\textbf{C}_3^{-1}\textbf{C}_2^\mathsf{T}
    \end{bmatrix}u^2    \label{eqn-31} \\
\begin{bmatrix}
   \textbf{b}_2-\textbf{C}_2^\mathsf{T}{\tilde{\textbf{X}}}
\end{bmatrix}
&=\begin{bmatrix}
   \textbf{C}_3
\end{bmatrix}{^W\tilde{\textbf{P}}_f}+\textbf{n}_2'' \label{eqn-32} \\
\textbf{R}_2''&=\begin{bmatrix}
   \textbf{C}_3
\end{bmatrix}u^2 \label{eqn-33}
\end{align}

Obviously, Eqs.~\eqref{eqn-30} and~\eqref{eqn-31} are equivalent residual equation and observation noise covariance. They could be substituted into standard EKF model Eqs.~\eqref{eqn-34} and~\eqref{eqn-37} to conduct state update directly.
\begin{align}
   \textbf{K}&=\textbf{PJ}^\mathsf{T}(\textbf{JPJ}^\mathsf{T}+\textbf{R})^\mathsf{-1}\label{eqn-34} \\
\Delta{\textbf{x}}&=\textbf{Kr}\label{eqn-35}  \\
\textbf{P}&\leftarrow(\textbf{I}-\textbf{KJ})\textbf{P}(\textbf{I}-\textbf{KJ})^\mathsf{T}+\textbf{KRK}^\mathsf{T}\label{eqn-36}  \\
\textbf{x}&\leftarrow\textbf{x}\oplus{\Delta{\textbf{x}}} \label{eqn-37}
\end{align}

\begin{table*}
   \small
   \centering
   \begin{threeparttable}
   \resizebox{\linewidth}{!}{
   \begin{tabular}{lcc|ccccccccccc}
     \toprule
     Sequence & S/M\tnote{1} & F/O\tnote{2} & MH1 & MH2 & MH3 & MH4 & MH5 & V11 & V12 & V13 & V21 & V22 & Avg \\
     \midrule
     OKVIS\tnote{4} \cite{okvis_2015keyframe}                  & M &   O & 0.160 & 0.220 & 0.240 & 0.340 & 0.470 & 0.090 & 0.200 & 0.240 & 0.130 & 0.160 & 0.225\\ 
     VINS-mono\cite{qin2018vinsmono}                           & M &   O & 0.150 & 0.150  & 0.220  & 0.320 & 0.300 & 0.079 & 0.110 & 0.180 & 0.080 & 0.160 & 0.174\\
     Kimera\cite{rosinol2020kimera}                            & S &   O & 0.110 & 0.100 & 0.160 & 0.240 & 0.350 & 0.050 & 0.080 & 0.070 & 0.080 & 0.100 & 0.134\\
     ICE-BA\cite{liu2018iceba}                                 & S &   O & 0.090 & 0.070 & 0.110 & 0.160 & 0.270 & 0.050 & 0.050 & 0.110 & 0.120 & 0.090 & 0.112\\
     SVO2.0\tnote{5}  \cite{Forster17troSVO}                   & S &   O & 0.080 & 0.080 & 0.088 & 0.211 & 0.231 & 0.052 & 0.082 & 0.073 & 0.084 & 0.116 & 0.109\\ 
     BASALT\cite{basalt2019}                                   & S &   O & 0.070 & 0.060 & \textbf{0.070} & 0.130 & 0.110 & 0.040 & 0.050 & 0.100 & 0.040 & 0.050 & 0.072\\
     DM-VIO\cite{dmvio2022}                                    & M &   O & 0.065 & \textbf{0.044} &  0.097 &  \textbf{0.102} &  \textbf{0.096} &  0.048 &  \textbf{0.045} &  0.069 &  \textbf{0.029} &  0.050 & \textbf{0.064}\\
     \midrule
     MSCKF\tnote{4}  \cite{mourikis2007msckf}                  & S &   F & 0.420 & 0.450 & 0.230 & 0.370 & 0.480 & 0.340 & 0.200 & 0.670 & 0.100 & 0.160 & 0.342 \\
     ROVIO\tnote{4}  \cite{rovio2017iterated}                  & M &   F & 0.210 & 0.250 & 0.250 & 0.490 & 0.520 & 0.100 & 0.100 & 0.140 & 0.120 & 0.140 & 0.232\\
     OpenVINS-4\tnote{5} \cite{geneva2020openvins}\tnote{3}    & S &   F & 0.084 & 0.084  & 0.127  & 0.218 & 0.360 & 0.038 & 0.054 & \underline{\textbf{0.050}} & 0.064 & 0.061 & 0.114 \\
     OpenVINS\tnote{5}  \cite{geneva2020openvins}              & S &   F & 0.072 & 0.143 & \underline{0.086} & 0.173 & 0.247 & 0.055 & 0.060 & 0.059 & 0.054 & \underline{\textbf{0.047}} & 0.096\\
     \textbf{SV(ours)}\tnote{5}                               & S &   F & \underline{\textbf{0.049}} & \underline{0.077} & \underline{0.086} & \underline{0.125} & \underline{0.125} & \underline{\textbf{0.035}} & \underline{0.053} & 0.082 & \underline{0.046} & 0.075 & \underline{0.075} \\

     \bottomrule
   \end{tabular}
    }

   \begin{tablenotes}
     \item[1] S and M mean stereo and monocular methods, respectively.
     \item[2] F and O mean filter-based and optimization-based methods, respectively.
     \item[3] OpenVINS-4 means that the maximum size of the sliding window in OpenVINS is configured to be 4.
     \item[4] results taken from \cite{delmerico2018benchmark}.
     \item[5] evaluated by author manually.
     \item[] All other results are taken from the respective paper.
   \end{tablenotes}
   \caption{Accuracy evaluation of various mono and stereo VINS algorithms on EuRoC. In the upper part, we summarize the results for the optimization-based methods that run sliding window optimization to estimate pose. In the lower part, we evaluate the results of filter-based methods. Best result in bold, underline is the best result among filter-based methods. SchurVINS achieves the lowest average APE RMSE in filter-based methods and surpasses the majority of optimization-based methods. }
   \label{tab:sota_rmse_euroc}
 \end{threeparttable}
 \end{table*}

\begin{table*}
   \small
   \centering
   \begin{threeparttable}
   \resizebox{\linewidth}{!}{
   \begin{tabular}{lcc|cccccccccccc}
     \toprule
     Sequence & S/M & F/O & c1 & c2 & c3 & c4 & c5 & r1 & r2 & r3 & r4 & r5 &r6 & Avg\\
     \midrule
     VINS-Mono\tnote{1}  & M &   O & 0.630  & 0.950 & 1.560 & 0.250 & 0.770 & 0.070 & \textbf{0.070} & 0.110 & 0.040 & 0.200 & 0.080  & 0.430  \\
     OKVIS\tnote{1}      & M &   O & 0.330 & 0.470 & 0.570 & 0.260 & 0.390 & 0.060 & 0.110 & 0.070 & 0.030 & 0.070 & 0.040 & 0.218  \\
     BASALT\tnote{1}     & S &   O & 0.340 & 0.420 & 0.350 & 0.210 & 0.370 & 0.090 & \textbf{0.070} & 0.130 & 0.050 & 0.130 & \textbf{0.020} & 0.198  \\
     DM-VIO\tnote{1}     & M &   O & \textbf{0.190} & 0.470 & \textbf{0.240} & \textbf{0.130} & \textbf{0.160} & \textbf{0.030} & 0.130 & 0.090 & 0.040 & 0.060 & \textbf{0.020} & \textbf{0.141}  \\
     \midrule
     ROVIO\tnote{1}      & M &   F & 0.470 & 0.750 & 0.850 & \underline{\textbf{0.130}} & 2.090 & 0.160 & 0.330 & 0.150 & 0.090 & 0.120 & 0.050 & 0.471  \\
     OpenVINS\tnote{2}   & S &   F & 0.413 & 0.322 & 1.536 & 0.186 & 0.644 & 0.062 & \underline{0.093} & 0.079 & \underline{\textbf{0.027}} & 0.074 & \underline{\textbf{0.020}}  & 0.314 \\
     \textbf{SV}\tnote{2}  & {S} & {F} & \underline{0.329} & \underline{\textbf{0.285}} & \underline{0.555} & {0.162} & \underline{0.274} & \underline{0.048} & {0.160} & \underline{\textbf{0.066}} & {0.049} & \underline{\textbf{0.054}} & {0.021} & \underline{0.182}  \\
     \bottomrule
   \end{tabular}
   }
   \begin{tablenotes}
      \item[1] results taken from \cite{dmvio2022}.
      \item[2] evaluated by author manually.
    \end{tablenotes}
   \caption{Accuracy evaluation on TUM-VI datasets. c1 to c5 denote corridor1 to corridor5 in TUM-VI datasets. r1 to r6 denote room1 to room6 in TUM-VI datasets. Best result in bold, underline is the best result among filter-based methods. }
   \label{tab:sota_rmse_tumvi}
 \end{threeparttable}
 \end{table*}

\subsection{EKF-based Landmark Solver}

$\tilde{\textbf{X}}$ can be obtained by substituting Eqs.~\eqref{eqn-30} and~\eqref{eqn-31} into Eqs.~\eqref{eqn-34} to~\eqref{eqn-37}. Then, the resulting $\tilde{\textbf{X}}$ could be substituted into~\cref{eqn-32} to establish the landmark equivalent residual equation
\begin{equation}
\begin{bmatrix}
   \textbf{r}_1 \\
   \textbf{r}_2 \\
   \vdots  \\
   \textbf{r}_m
\end{bmatrix} = 
\begin{bmatrix}
   \textbf{C}_{3_1} & & & \\
   & \textbf{C}_{3_2} & & \\
   & & \ddots  & \\
   & & & \textbf{C}_{3_m}
\end{bmatrix}
\begin{bmatrix}
   ^W\tilde{\textbf{P}}_{f_1} \\
   ^W\tilde{\textbf{P}}_{f_2} \\
   \vdots  \\
   ^W\tilde{\textbf{P}}_{f_m}
\end{bmatrix} + \textbf{n}_2''       \label{eqn-38}
\end{equation}
where $\textbf{C}_{3_1}, \dotsb, \textbf{C}_{3_m}$ are diagonal elements of $\textbf{C}_3$ clarified in~\cref{eqn-23}. And the corresponding covariance $\textbf{R}_2''$ is:
\begin{equation}
   \textbf{R}_2''=\begin{bmatrix}
      \textbf{C}_{3_1}u^2 & & & \\
   & \textbf{C}_{3_2}u^2 & & \\
   & & \ddots  & \\
   & & & \textbf{C}_{3_m}u^2
\end{bmatrix}    \label{eqn-39}
\end{equation}

Benefited from the sparsity of the resulting landmark equivalent residual equation, Eqs.~\eqref{eqn-38} and~\eqref{eqn-39} is split as a bunch of small independent residual models, shown as~\cref{eqn-40}, which allows the EKF update of each landmark to conduct one by one. This significantly reduces the computational complexity.
\begin{equation}
\begin{aligned}
\begin{bmatrix}
   \textbf{r}_i
\end{bmatrix} &= 
\begin{bmatrix}
   \textbf{C}_{3_i}
\end{bmatrix}
\begin{bmatrix}
   ^W\tilde{\textbf{P}}_{f_i}
\end{bmatrix} + \textbf{n}_{2_i}'', i=1,\dotsb,m  \\
\textbf{R}&=[\textbf{C}_{3_i}u^2]
\end{aligned}
\label{eqn-40}
\end{equation}


\subsection{Frontend}
Our code implementation makes full use of SVO2.0 as the front-end of SchurVINS. The integrated components of SchurVINS include feature alignment and depth-filter modules from original SVO2.0. Meanwhile, sparse image alignment module is replaced by the proposed EKF propagation scheme to guarantee delivering an accurate pose to feature alignment module. Compared with frame-to-frame feature tracking\cite{qin2018vinsmono,sun2018smsckf,geneva2020openvins}, the strategy of feature alignment, implemented by projecting and matching the co-visible landmarks from local map to frames, achieves excellent long-term landmark tracking performance due to the fact that the lost landmarks in short time is capable to be tracked again. Depth-filter is utilized to execute landmark position initialization. Once the landmark is initialized sufficiently, it would be transferred to the proposed EKF-based landmark solver to proceed estimation with sliding window jointly.

Based on First In First Out (FIFO) strategy, local map only maintains the most recent ten keyframes to support landmark tracking. Since high accuracy is already achieved, the traditional LBA is no longer necessary, which is abandoned in the proposed SchurVINS. 

\begin{center}
   \begin{table}[t]
     \centering
     \begin{threeparttable}
         \scriptsize
        \resizebox{\linewidth}{!}{
        \begin{tabular}{l|p{1.8cm}p{1.8cm}p{1.8cm}}
           \toprule
           & Avg CPU & Std CPU & Speed \\
           \midrule
           DM-VIO        & 98/172\tnote{1} & -/30 & 1x/1.76x \\
           BASALT        & 46/203\tnote{1} & -/46 & 1x/4.37x \\
           VINS-Mono        & 45 & 13 & 1x \\
           OpenVINS        & 37 & 10 & 1x \\
           OpenVINS-4        & 32 & 8 & 1x \\
           SMSCKF\cite{sun2018smsckf}        & 25 & 4 & 1x \\
           SVO2.0        & 89 & 20 & 1x \\
           SVO2.0-wo\tnote{2}  & 17 & 6 & 1x \\
           SV  & {18} & 6 & {1x} \\
           \bottomrule
        \end{tabular}
        }
        \begin{tablenotes}
           \item[1] The 1x evaluation results of DM-VIO and BASALT are the converted results by\\ author manually.
           \item[2] SVO2.0-wo means SVO2.0 without the enabled LBA.
        \end{tablenotes}

        \caption{Evaluation of CPU overhead for different wellknown VINS algorithms. GBA, PGO and LC are disabled on all the mentioned algorithms, with the exception of SVO2.0, which has the LBA module enabled. Our method provides a notable improvement in efficiency compared to the SOTA VINS algorithms.}
        \label{tab:cpu_profile}
     \end{threeparttable}
   \end{table}
   \end{center}
   
\subsection{Keyframe Selection}

The strategy of keyframe selection is important in VINS system. There are three strategies to select keyframes in SchurVINS. If the average parallax between the candidate frame and the previous keyframe reaches the threshold or the count of tracked landmarks drops below the certain threshold, the corresponding frame is defined as keyframe. Once the keyframe is selected, the FAST corners~\cite{trajkovic1998fast} are extracted to generate new landmarks via depth-filter module. Additionally, when the gap in both orientation and position between the candidate frame and the co-visible keyframes maintained in the local map is out of the certain range, the keyframe would be determined, by which the tracking module could overcome divergence between the candidate frame and the local map.

%% file: sec/4_experiment.tex
\section{Experiments}
\label{sec:Experiments}

The accuracy and efficiency of SchurVINS algorithms are evaluated by two experiments. And the additional ablation experiment is carried out to demonstrate the effectiveness of the proposed EKF-based landmark solver.

\textbf{System Configuration:} We have developed SchurVINS based on the open source code repository of SVO2.0, specifically, svo\_pro\_open. The majority of system parameters are not required to be modified. For high efficiency, edgelet features, loop closure (LC), pose graph optimization (PGO), LBA and Global BA (GBA) are discarded or deactivated. For our experiments below, we have configured the threshold on the quantity of keyframes in the local map to a maximum of ten. This local map mainly maintains co-visible keyframes and landmarks to achieve feature alignment. In the backend of SchurVINS, there is a sliding window consists of 2 old keyframes and 2 latest temporal frames. The keyframe strategy is similar to original SVO2.0.

\begin{table}[t]
  \begin{threeparttable}
     \scriptsize 
     \centering
     \resizebox{\linewidth}{!}{
      \begin{tabular}{l|p{0.4cm}p{0.4cm}p{0.4cm}llll}
       \toprule
        &
       \begin{sideways}
         {SV}
       \end{sideways}
        & 
        \begin{sideways}
         SV-GN\tnote{1}
       \end{sideways}
       & 
       \begin{sideways}
         SVO2.0-wo
       \end{sideways}
       & 
       \begin{sideways}
         SVO2.0
       \end{sideways}
       & 
       \begin{sideways}
        SMSCKF\cite{sun2018smsckf}
      \end{sideways}
      & 
       \begin{sideways}
         OpenVINS
       \end{sideways}
       & 
       \begin{sideways}
         OpenVINS-4
       \end{sideways}
       \\
       \midrule
       SparseImageAlign   & {-} &   - & 1.35 & 1.43 &-& - & -  \\
       FeatureAlign  & {1.39} & 1.39 & 1.79 & 1.91  &-& -  & -   \\
       KLT                & {-} & - & - & - &2.63& 2.69 & 2.67  \\
       Propagation        & {0.11} & 0.11 & - & - &0.55& 0.21 & 0.18  \\
       optimizePose       & {0.67} & 0.67 & 0.48 & - &3.16& 0.99/4.30\tnote{2}  & 0.34/2.46\tnote{2}  \\
       optimizeStructure  & {0.11} & 0.42 & 0.07 & - &-& 0.93  & 0.44  \\
       LBA                & {-} & - & - & 26.3\tnote{3} &-& - & -  \\
       \midrule
       Total time\tnote{4}         & {3.83} & 4.11 & 3.77 & 9.28 &8.53& 10.91 & 7.89  \\
       \bottomrule
     \end{tabular}
     }
     \begin{tablenotes}
       \item[1] denotes SchurVINS with Gauss-Newton optimization-based (GN-based) land-\\mark estimation as originally used in SVO2.0.
       \item[2] Running time of MSCKF update and SLAM update.
       \item[3] It contains some running time of SVO2.0 LBA in asynchronous thread.
       \item[4] The total time also contains other modules.
     \end{tablenotes}
     \caption{Running time evaluation of the main parts of SchurVINS compared with SVO2.0 and OpenVINS on EuRoC MH01 (mean time in ms). Note that the different overhead of optimizeStructure between SVO-NonBA and SchurVINS-GN is primarily attributed to the variation in the count of feature matches, which is a consequence of the localization accuracy.}
     \label{tab:compare_runtime}
  \end{threeparttable}
\end{table}

\subsection{Accuracy} 
The overall accuracy of the mentioned algorithms is evaluated using Root Mean Square Error (RMSE) on two wellknown datasets, EuRoC~\cite{burri2016eurocdata} and TUM-VI~\cite{tum_dataset2018tum}. The corresponding experimental trajectory and point cloud of SchurVINS on TUM-VI and EuRoC datasets are shown on~\cref{fig:trajectory_and_points}. To prevent the fluctuation of the algorithm from causing unreasonable evaluation results, our own evaluation method is to run the algorithm for 7 rounds, remove the maximum and minimum values, and then calculate the average of the remaining results as the evaluation result. 
\begin{table*}[t]
  \begin{threeparttable}
  \small
  \centering
  \resizebox{\linewidth}{!}{
  \begin{tabular}{l|p{1.2cm}p{1.2cm}p{1.2cm}p{1.2cm}p{1.2cm}p{1.2cm}p{1.2cm}p{1.2cm}p{1.2cm}p{1.2cm}p{1.2cm}}
    \toprule
    Sequence & MH1 & MH2 & MH3 & MH4 & MH5 &   V11 &  V12 &  V13 &  V21 &  V22 &  Avg \\
    \midrule
    SV               &   \textbf{0.049} & 0.077         & \textbf{0.086} & \textbf{0.125} & 0.125         & \textbf{0.035} & 0.053         & 0.082         & 0.046         & \textbf{0.075} & 0.075 \\
    SV-GN            &   0.057         & \textbf{0.055} & 0.097         & 0.135         & \textbf{0.116} & 0.038         & \textbf{0.051} & \textbf{0.068} & \textbf{0.037} & 0.083         & \textbf{0.073} \\
    SV-OFF\tnote{1}  &   0.067        &	0.103         & 0.107         &	0.137         &	0.143         & 0.038         &	0.062         &	-             &	0.057         &	0.255         & 0.107 \\
    \bottomrule
  \end{tabular}
  }
  \begin{tablenotes}
   \item[1] SV-OFF denotes SchurVINS with disabled EKF-based landmark solver only uses depth-filter to initialize landmark.
  \end{tablenotes}
  \caption{Ablation Evaluation on EuRoC.}
  \label{tab:ablation_table}
\end{threeparttable}
\end{table*}
In \cref{tab:sota_rmse_euroc}, our method obtains the lowest average RMSE in filter-based methods reported on the dataset so far, as well as outperforms the majority of optimization-based methods. Besides, our approach obtains the similar accuracy with wellknown optimization-based method BASALT and slightly lower accuracy than the recent competitor DM-VIO. Besides, the well-known VINS algorithms, VINS-Fusion~\cite{qin2018vinsmono} and SMSCKF~\cite{sun2018smsckf}, are not included in~\cref{tab:sota_rmse_euroc}, since VINS-mono and OpenVINS surpass VINS-Fusion and SMSCKF in terms of accuracy, respectively~\cite{vinsfusion,geneva2020openvins}. The re-evaluation experiment in~\cref{tab:sota_rmse_tumvi} is within expectation absolutely. It is worth highlighting that, although degrading in accuracy slightly compared with the two optimization-based competitors, our method achieves obviously lower computational complexity than both of them with details in the next subsection.

\subsection{Efficiency}
The efficiency evaluations are carried out on Intel i7-9700 (3.00GHZ) desktop platform. Global BA (GBA), pose graph optimization and loop closure are disabled on all of the following algorithms. Besides, LBA is only enabled on the original SVO2.0. The efficiency experiment is divided into two parts: profiling processor usage and overhead time, which are reported in \cref{tab:cpu_profile} and \cref{tab:compare_runtime}, respectively.

As demonstrated in \cref{tab:cpu_profile}, SchurVINS achieves almost the lowest processor usage compared with all the mentioned VINS algorithms. Especially, SVO2.0-wo requires similar cpu usage with SchurVINS, but it suffers from notable inaccuracy since it is almost pure Visual Odometry (VO). To thoroughly investigate the underlying reasons contributing to the efficiency advantages of SchurVINS, the experiment to meticulously analyze the overhead time of SchurVINS including the comparison with SVO2.0, the widely-recognized filter-based OpenVINS and SMSCKF is carried out in \cref{tab:compare_runtime}.

In \cref{tab:compare_runtime}, the optimizeStructure module in SchurVINS is nearly 3 times faster than that of SchurVINS-GN. Because our method obtains significant computational savings by leveraging the intermediate results of Schur complement. In contrast, SchurVINS-GN reconstructs problems to estimate landmarks. Compared with SVO2.0-wo, SchurVINS is faster due to its replacement from the high-computational SparseImageAlign to propagation module. In contrast, the optimizeStructure of SVO2.0-wo is obviously faster than SchurVINS-GN. The reason is that the latter utilizes almost 4 times measurements than the former to conduct optimization. Compared with SVO2.0, the root cause leads to the obviously increased run time of algorithm is the high computational complexity of LBA. In consideration of OpenVINS, it is noteworthy that neither the default configuration nor the configuration with a maximum size of sliding window of 4 could achieve that OpenVINS outperforms SchurVINS in efficiency. What stands out from this analysis is that the update of SLAM points in OpenVINS requires noticeably more computational resources compared with the EKF-based landmark estimation presented in SchurVINS. Illustrated on \cref{fig:CovAndHessianCompareFig2}, SchurVINS makes full use of the sparsity of problem than both hybrid MSCKF and optimization-based methods.
\subsection{Ablation Study}
The experiments above strongly support SchurVINS. And thus it is necessary to study the impact of different components of our algorithm. Based on SchurVINS, we replace or discard the EKF-based landmark solver to analyse its effectiveness.

As illustrated in \cref{tab:ablation_table}, if without either GN-based or EKF-based  landmark solver, SchurVINS cannot sufficiently limit the global drift. Moreover, in some challenge scenarios, lack of estimating landmarks simultaneously in SchurVINS may lead to system divergency. The comparison between SchurVINS and SchurVINS-GN in \cref{tab:ablation_table} indicates that both the proposed EKF-based landmark solver and the GN-based landmark solver belonging to original SVO2.0 are effective and reliable to guarantee high precision. In addition, the comparison between them in \cref{tab:compare_runtime} and \cref{tab:ablation_table}, illustrates that although the proposed EKF-based landmark solver leads to slight accuracy degradation, it could achieve the obviously low computational complexity. An intuitive explanation for the decreased accuracy is that our method only uses all the observations in sliding window for landmark estimation.

%% file: sec/5_conclusion.tex
\section{Conclusions and Future Work}
\label{sec:Conclusion}

In this paper, we have developed an EKF-based VINS algorithm, including the novel EKF-based landmark solver, to achieve 6-DoF estimation with both high efficiency and accuracy. In particular, the formulated equivalent residual model consisting of Hessian, Gradient and the corresponding observation covariance is utilized to estimate poses and landmarks jointly to guarantee high-precision positioning. To achieve high efficiency, the equivalent residual model is decomposed as pose residual model and landmark residual model by Schur complement to conduct EKF update respectively. Benefited from the probabilistic independence of surrounding environment elements, the resulting landmark residual model are split as a bunch of small independent residual models for the EKF update of each landmark, which significantly reduces the computational complexity. To best of our knowledge, we are the first to utilize Schur complement factorizing residual model in EKF-based VINS algorithms for acceleration. The experiments based on EuRoC and TUM-VI datasets demonstrate that our approach notably outperforms the overall EKF-based methods~\cite{sun2018smsckf,geneva2020openvins} and the majority of optimization-based methods in both accuracy and efficiency. Besides, our approach requires almost less than 50$\%$ computational resource than the SOTA optimization-based methods~\cite{basalt2019,dmvio2022} with comparable accuracy. In the meanwhile, the ablation studies clearly demonstrate that our proposed EKF-based landmark solver is not only significantly efficient but also capable of ensuring high accuracy.

In future work, we will focus on the local map refinement in SchurVINS to explore more accuracy.

%% file: sec/6_acknowledgment.tex
\section{Acknowledgment}
\label{sec:Acknowledgment}
We would like to thank Taoran Chen, Chen Chen, and Jiatong Li in ByteDance as well as Zihuan Cheng in SCUT for their kind help. Moreover, I (Frank) would like to deeply thank my wife, Linan Guo.